\pdfoutput=1
\documentclass[preprint,journal]{vgtc}       

\ifpdf
  \pdfoutput=1\relax                   
  \usepackage{graphicx}                
  \DeclareGraphicsExtensions{.pdf,.png,.jpg,.jpeg} 
\else
  \usepackage{graphicx}                
  \DeclareGraphicsExtensions{.eps}     
\fi%

\usepackage{cite}
\usepackage{amsmath,amssymb,amsfonts}
\usepackage{algorithmic}
\usepackage{graphicx}
\usepackage{textcomp}

\usepackage{balance}  

\usepackage{url}      
\usepackage[lined, boxed, figure]{algorithm2e}
\usepackage{enumitem}
\usepackage{xspace}
\usepackage[table,usenames,dvipsnames]{xcolor}
\usepackage{booktabs}
\usepackage{multirow}
\usepackage[10pt]{moresize} 
\usepackage{changepage}
\usepackage{soul}
\usepackage{float}
\usepackage{caption}

\newcommand{\name}{EcoLens\xspace}

\definecolor{blue}{RGB}{31,119,180}
\definecolor{orange}{RGB}{255, 127, 14}
\definecolor{green}{RGB}{44, 160, 44}
\definecolor{visblue}{RGB}{51,105,204}
\definecolor{visorange}{RGB}{255, 153, 0}
\definecolor{black}{RGB}{0, 0, 0}

\makeatletter
\def\itemautorefname{\@gobble}
\makeatother

\renewcommand{\manuscriptnotetxt}{}
\nocopyrightspace

\vgtccategory{Research}
\vgtcpapertype{algorithm/technique}

\title{\name: Visual Analysis of Urban Region Dynamics Using Traffic Data}

\author{Zhuochen Jin, Nan Cao, Yang Shi, Hanghang Tong, Yingcai Wu}
\authorfooter{
\item
Intelligent Big Data Visualization Lab, Tongji University, Shanghai, China
\item
School of Computing Informatics and Decision Systems Engineering, Arizona State University, USA
\item
College of Computer Science and Technology, Zhejiang University, Hangzhou, Zhejiang, China
}

\abstract{The rapid development of urbanization during the past decades has significantly improved people's lives but also introduced new challenges on effective functional urban planning and transportation management. The functional regions defined based on a static boundary rarely reflect an individual's daily experience of the space in which they live and visit for a variety of purposes. Fortunately, the increasing availability of spatiotemporal data provides unprecedented opportunities for understanding the structure of an urban area in terms of people's activity pattern and how they form the latent regions over time. These ecological regions, where people temporarily share a similar moving behavior during a short period of time, could provide insights into urban planning and smart-city services. However, existing solutions are limited in their capacity of capturing the evolutionary patterns of dynamic latent regions within urban context. In this work, we introduce an interactive visual analysis approach, EcoLens, that allows analysts to progressively explore and analyze the complex dynamic segmentation patterns of a city using traffic data. We propose an extended non-negative Matrix Factorization based algorithm smoothed over both spatial and temporal dimensions to capture the spatiotemporal dynamics of the city. The algorithm also ensures the orthogonality of its result to facilitate the interpretation of different patterns. A suite of visualizations is designed to illustrate the dynamics of city segmentation and the corresponding interactions are added to support the exploration of the segmentation patterns over time. 
We evaluate the effectiveness of our system via case studies using a real-world dataset and a qualitative interview with the domain expert.
} 


\usepackage{color}

\begin{document}

\maketitle

\section{Introduction}
\label{sec:intro}
The rapid development of urbanization during the past decades has significantly improved people's life, but also introduced new challenges on effective functional urban planning and transportation management to serve various needs for its citizens. 
Tremendous efforts have been put on reasonably and optimally segmenting the city into functional regions to best utilize the limited city resource. 
Government agencies typically define functional regions based on a static boundary.
However, the segmentation rarely reflects an individual's day-to-day experience of the space in which they live and visit for a variety of purposes. The use of these static boundaries has limited not only the city's ability to assess the dynamic processes that shape the city's urban areas, but also its opportunity to improve the city management with smart-city services ~\cite{su2011smart, bakici2013smart}.
Therefore, a better understanding of the structure of an urban area in terms of people's activity patterns and how they form the latent regions over time will provide profound insights for valuable applications in urban planning and business intelligence.
The increasing availability of human mobility data generated within urban context opens up unprecedented opportunities for us to engage in data-driven science and better understand an urban area. 
In this work, we propose an analysis technique to explore the {\it ecological regions} (i.e., {\it dynamic latent regions}) in the urban context that reflect the city dwellers' dynamic moving patterns and capture how they share similar moving behavior during a short period of time.

There has been prior work studying mobility patterns in the urban context~\cite{zheng2016visual}, and most of them focus on identifying predefined events or features in data (e.g.~\cite{ferreira2013visual, wu2016telcovis, zheng2015visual, zheng2016telcoflow, zeng2013visualizing}), which cannot capture the dynamic formation of regions in an urban area. There have been some attempts in developing automatic algorithms~\cite{wang2014discovering, yuan2015discovering} to extract latent regions in an urban area. However, analysts found it difficult to understand and interpret the result or combine their domain knowledge with real-world applications. Hence, there exists a need to involve human supervision in the analysis of dynamic latent regions. 

Visual analysis provides an effective way to integrate humans in a data exploration process by applying their perceptual abilities to the target dataset and leveraging their domain knowledge to guide the exploration. The state-of-the-art approach, MobiSeg~\cite{wu2016mobiseg}, enables an interactive exploration of people's movement to segment an urban area into regions while neglecting the continuity of people's patterns in either spatial or temporal domain. 
Thus, its approach cannot be directly employed to illustrate the evolutionary patterns of dynamic latent regions in the urban context. 
To address this issue, we introduce an interactive visual analysis approach, \textit{\name}, which allows analysts to progressively explore and analyze the complex evolutionary patterns of latent regions. In particular, we introduce a novel non-negative matrix factorization based (NMF-based) algorithm for dynamic latent region detection based on people's mobility patterns, which takes the temporal and spatial smoothness into consideration. Meanwhile, a suite of visualizations is designed to illustrate the extracted regions within spatial and temporal context. The major contributions of this paper are summarized as follows:
\begin{itemize}[noitemsep]

\item \textbf{System.}
A novel visual analytic system is presented to dispose the large scale of traffic data and explore the dynamic segmentation patterns of a city. The visualization views are specifically designed to illustrate the temporal changing trend of city segmentations and the corresponding interactions are designed to support an easy exploration of the segmentation patterns over time.

\item \textbf{Algorithm.}
We introduce a novel NMF-based algorithm smoothed over both spatial and temporal dimensions to capture the spatiotemporal dynamics of the city. Orthogonality of the latent patterns is guaranteed to facilitate the interpretation of different patterns.

\item \textbf{Evaluation.}
We demonstrate the performance of \name through case studies using over 450,000 taxi trips collected in Manhattan from July 2014 to December 2014 which is a large-scale real-world dataset. We also report the feedback from an expert in the field of urban planning.
\end{itemize}

The rest of the paper is organized as follows: after a brief survey of related work in Section 2, we present an overview of the proposed approach and our system in Section 3. Then in Section 4, we describe the details of the algorithm for dynamic latent region analysis, followed by the introduction of visualization designs and interaction techniques employed in our system in Section 5. To evaluate the system, Section 6 presents case studies based on the real-world data collected from Yellow Cabs in Manhattan area of New York City as well as expert feedback. Finally, we summarize the paper in Section 7.

\section{Related Work}
\label{sec:related}
We review the prior works that are relevant to this paper, including (1) urban segmentation and (2) visualization of mobility patterns.

\subsection{Urban Segmentation}
Urban segmentation has been studied extensively for years in the fields of urban planning and geographic information system (GIS). Remote sensing data (e.g., images) are frequently used ~\cite{deng2009spatio,seto2005quantifying,puissant2005utility}. These techniques recognized different regions by calculating visual differences based on satellite images. The results are usually limited by the low resolution and the missing of context details especially in regions with complicated geographic conditions. To address this issue, more and more research attention has been put on identifying functional urban regions based on people's daily activity patterns~\cite{jiang2012discovering, pan2013land, wakamiya2011crowd, wang2014discovering, wu2016mobiseg, yuan2015discovering}. As early as the 1970s, Goddard et al.~\cite{goddard1970functional} analytically differentiated functional regions in central London based on taxi flows. Following this work, Yuan et al.~\cite{yuan2015discovering} recently employs Latent Dirichlet Allocation (LDA), a generative statistical model that was originally designed for text analysis, to identify the latent functional regions in a city based on people's mobility patterns. Kraft et al. ~\cite{kraft2017delimitation} applied the local minimum and maximum values of transport intensities to delimit functional regions. Dem{\v{s}}ar et al.~\cite{demvsar2017revisiting} applied principal components analysis (PCA) to taxi flows for obtaining functional regions. Zhang et al.~\cite{zhang2017analyzing} analyzed the patterns of the urban roads based on taxi GPS data. Wakamiya et al.~\cite{wakamiya2015twitter} applied non-negative matrix factorization (NMF) to analyze urban area characterization based on Twitter data. When compared to these techniques which produce static segmentation results, our work focuses on revealing the regional dynamics. The algorithms and visualization designs are thus introduced.

Despite aforementioned analysis driven methods, MobiSeg~\cite{wu2016mobiseg} is the first visual analysis system designed for interactive region segmentation in the urban context. As the most relevant work, MobiSeg also employed NMF for urban segmentation and introduced the visual interface to facilitate results interpretation and interactive latent region analysis, comparison, and exploration. When compared to MobiSeg, our work focuses on analyzing and revealing temporal patterns of the dynamic transition of the latent regions. To this end, we introduce a dynamic NMF algorithm that analyzes and smooths the transition over both temporal and spatial domain, which produces more continuous and interpretable results when compared to the results without smoothing. In addition, to facilitate the tracking of regional transition patterns and the change of the regional functions, we also introduce a pattern tracking algorithm based on a Sankey diagram design. Most importantly, we conduct case studies on real data which reveal interesting findings that can hardly be detected in MobiSeg.

\subsection{Visualization of Mobility Patterns}
Visual analysis of latent urban regions falls into the general topic of analyzing mobility patterns. Efforts have been devoted to developing visualization methods to meet the needs of analyzing and understanding mobility patterns within the urban context (see ~\cite{zheng2016visual} for a comprehensive survey). Our goal in this work is to develop a visual analysis approach to present the evolution of mobility patterns, thus helping analysts explore and understand dynamic latent regions within the urban context. When presenting evolving mobility patterns, time and movement are two fundamental components of telling a full story and can help structure the information. Therefore, in this part, among the vast amount of visualization techniques, we focus on visualization of time and movement which are the most relevant to our work.

There are various ways of mapping time to visual variables~\cite{aigner2011visualization}. Within an urban context, the axis-based design is one of the most popular methods~\cite{ferreira2013visual, wu2014boundaryseer, zheng2016telcoflow} due to its simplicity and interpretability. Besides, temporal information can also be conveyed through a dynamic representation, resulting in visualizations that change over time automatically (i.e., animation), which is a popular design choice for visualizing the dynamics of a city~\cite{kloeckl2011live, rosling2009gapminder}.
However, as demonstrated by Robertson et al.~\cite{robertson2008effectiveness}, the animation techniques are generally not effective for analysis tasks due to the limitation of human short-term memory. We employ the both designs in our work. A Sankey diagram is used to provide an overview of the regional transition trend and an animated map view to illustrate the dynamic change of regions in the spatial context.

When visualizing movements, there are three major types of techniques, including direct depiction, summarization, and pattern extraction techniques~\cite{andrienko2013visual}. Direct depiction techniques~\cite{guo2011tripvista, tominski2012stacking} present paths of movement directly. Summarization techniques~\cite{guo2014origin, wu2016telcovis, andrienko2017revealing} conduct statistical calculations of movement and present the result based on divided spatial or temporal intervals. Pattern extraction techniques~\cite{zheng2015visual, zheng2016telcoflow} enable an interactive discovery and analysis of various movement patterns. In this paper, we integrate different types of techniques and enhance them with new features. With \name, analysts could observe the evolution of an area frame by frame and explore its corresponding mobility patterns interactively. 
\section{System Overview}
\label{sec:overview}

\begin{figure*}[!t]
\centering
\includegraphics[width=\linewidth]{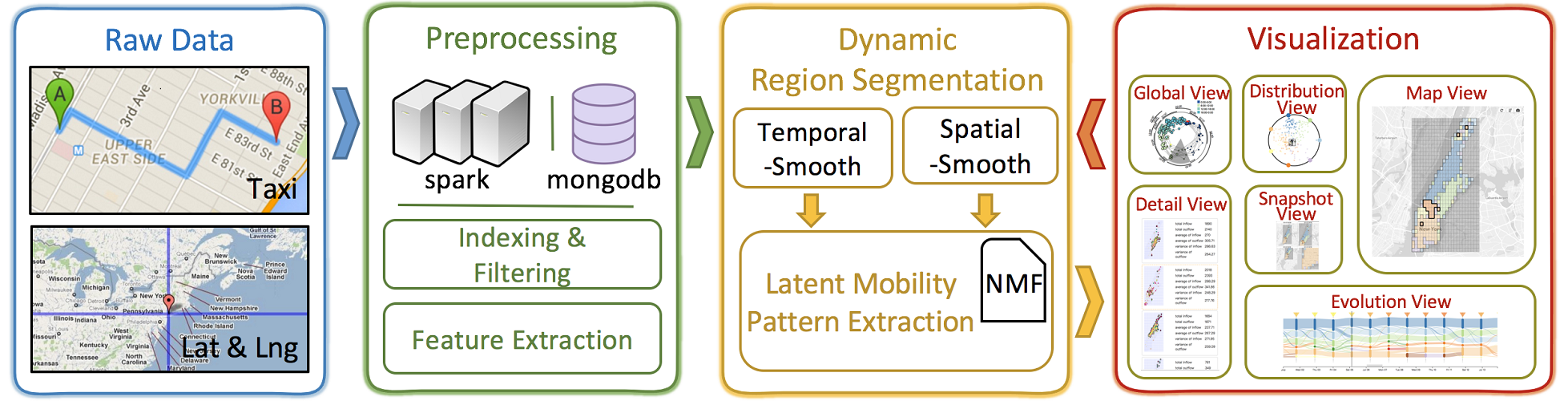}
\caption{The system pipeline of \name. Three primary modules, including preprocessing module, analysis module, and visualization module, support the analysis of evolution patterns of latent regions within the urban context.}
\label{fig:system}
\end{figure*}

\name is designed for revealing the dynamics of latent regions in which people share similar temporal mobility patterns.
Following a user-centric design process, we worked closely with a domain expert who is a researcher at the institute of urban planning and design in China. Regular meetings with the expert were scheduled and lasted for about six months to help us understand the requirements and refine the prototype. During the meetings, we focused on discussing what kinds of dynamic patterns need to be captured and how to reveal and interpret them within the urban context. 
As a result, we believe people's mobility pattern is the most critical feature to capture as it is directly relevant to people's daily behaviors or habits, which may imply the functionality of a region. 
For example, during the weekdays, people tend to travel from home to the office in the morning. This mobility pattern suggests potential residential areas and business areas.
More specifically, the desired system should satisfy the following requirements: 
\begin{enumerate}
\item[{\bf R1}] {\bf Identifying the mobility patterns occurred in different areas to form latent regions for investigation.} The system should be able to differentiate mobility patterns occurred in various urban areas based on people's collective daily moving behaviors that are extracted from a large dataset. Areas with similar patterns should be further grouped into latent regions to help imply the corresponding regional functionality in urban context.
\item[{\bf R2}] {\bf Capturing the dynamics of the latent regions over time to uncover regular or irregular regional transition patterns.} The system should be able to reveal the dynamic change of the spatial mobility patterns and the corresponding change of the latent regions. This will help analysts identify the regular changes due to people's regular daily behaviors and thus help them identify those irregular ones due to certain events.
\item[{\bf R3}] {\bf Facilitating the regional pattern comparison, exploration, inspection, and interpretation.} The system should also be able to intuitively illustrate the aforementioned patterns and the corresponding changes of the latent regions in the rich urban context so that analysts can easily explore, compare, and understand the patterns and their changes to make a proper conclusion as well as a correct decision.
\end{enumerate}
According to the design requirements, we develop \name, an interactive visualization system for analyzing the evolution patterns of latent regions within the urban context. The system consists of three primary modules (Fig.~\ref{fig:system}): 
(1) The \textit{prepossessing module} is designed to clean the raw data (i.e., the taxi-trips in our case) and transform them into the matrix time series with desired features. This whole prepossessing step runs in parallel on a spark cluster and the processed data are stored in MongoDB\footnote{https://www.mongodb.com/} for later querying. This module is designed to extract the collective moving behaviors of people from raw data (\textbf{R1}). (2) The \textit{analysis module} derives the latent regions (\textbf{R1}) and the evolution patterns (\textbf{R2}) over time based on the preprocessed data via non-negative matrix factorization. (3) The \textit{visualization module} presents the analysis results via multiple coordinated views. These views reveal the evolution of the latent regions and facilitate the interpretation of the corresponding pattern within each region. Various interactions are provided to support flexible data exploration and result calibration (\textbf{R3}). In the next, we will describe the details of the analysis and visualization modules in the following sections.
\section{Context Preserving Dynamic Region Segmentation}
\label{sec:algorithm}
In this section, we introduce the algorithm used in the analysis module that is developed for revealing the regional dynamics of the traffic evolution patterns inside a focal urban area. The proposed algorithm leverages the non-negative matrix factorization (NMF)~\cite{lee1999learning} and smooths the change over both the spatial and the temporal dimensions to facilitate interpretation. The rest of the section will first describe the data and the corresponding features used in our prototype system, followed by the algorithm details as well as the design rationales.

\subsection{Feature Extraction}
Our prototype system employs the public New York City taxi trip dataset~\footnote{\url{http://www.nyc.gov/html/tlc/html/about/trip\_record\_data.shtml}} to capture the change of regional mobility patterns. To this end, we divided the Manhattan area, our focal investigation region, into $N$ grids ($N = 300$ with the granularity of $0.005 \text{ longitude}\times 0.005 \text{ latitude}$ in our implementation) and counted the number of incoming and outgoing trips in each grid as the grid's features for later analysis. Each grid $i$ is described by a $2N$-dimensional feature vector with the field $p$ in the vector indicates the number of trips from the $i$-th grid to the $p$-th grid. Therefore, the first $N$ fields in the vector indicate the number of outgoing trips from grid $i$ to the rest $N-1$ grids and the last $N$ fields indicate the number of incoming trips from other grids to the $i$-th grid.

In this way, during a given time interval $t$ ($t = 2$ hours in our implementation),  a $2N\times N$ feature matrix $\mathbf{X}_{t}$ can be obtained, which captures the mobility patterns during $t$. The matrices from different intervals thus formed a feature matrix time series. This matrix series characterizes mobility patterns in each region over time and are used for the latter analysis.

\begin{figure}[htbp]
    \centering
    \includegraphics[width=\linewidth]{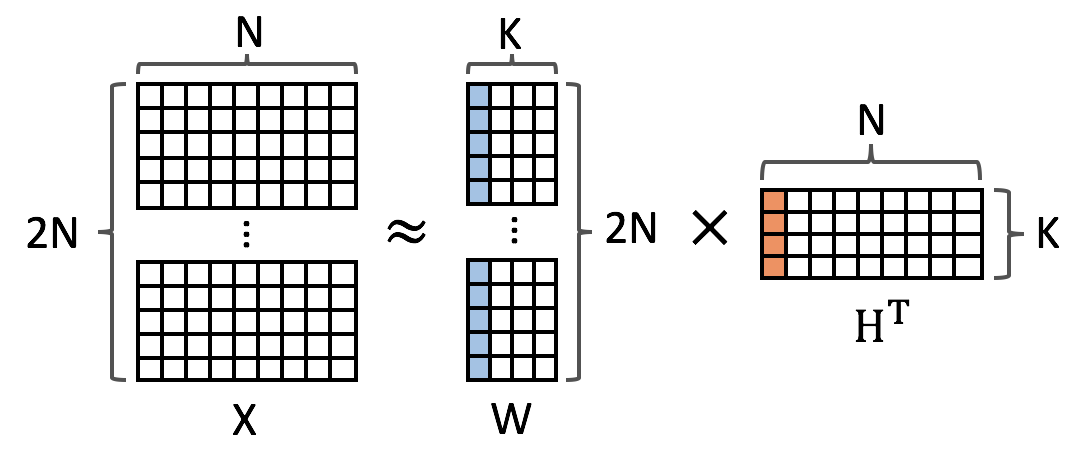}
    \caption{Matrix $X$ is decomposed into the product of matrix $H$ and $W$. The column in $H^{T}$ highlighted in red indicates the likelihood of the $K$ mobility patterns occurred in the $1^{st}$ regions.
    The column in $W$ highlighted in blue shows the probability of the $1^{st}$ mobility pattern having a certain feature.}
	\label{fig:matrix}
\end{figure}

\subsection{Dynamic Region Segmentation} 
We propose a context preserving algorithm based on non-negative matrix factorization (NMF) to analyze the regional dynamics of the taxi trips captured in the aforementioned feature matrix time series. The algorithm optimizes and balances among four carefully designed terms that are formally defined as follows:
\vspace{-0.1cm}
\begin{align}
W_t,H_t,M_t = 
    &\mathop{argmin}(||X_t - W_t H_t^T||^2 \\
    & + \alpha ||W_{t-1} - W_t M_t^T||^2 \\
    & + \beta ||H_{t-1} - H_t M_t^T ||^2 \\
    & + \lambda \sum_{i,j} A_{i j} ||h_{t i} -h_{t j} ||^2 )
\label{eq:OP} 
\end{align}
subject to 
\[
W_t^T W_t =I, t=1,2,\ldots,n, W_t \succeq 0, H_t \succeq 0, M_t \succeq 0
\]

\textbf{Pattern Extraction.} The first term is proposed to extract latent mobility patterns from the raw data. It employs the non-negative matrix factorization (NMF) to decompose a feature matrix $(\mathbf{X}_{t})_{2N \times N}$ into the product of two non-negative matrices $(H_{t})_{N \times K}$ and $(W_{t})_{2N \times K}$ that respectively captures the spatial distribution of the latent patterns as well as the pattern semantics, as shown in Fig~\ref{fig:matrix}. Specifically, $(H_{t})_{N \times K}$ indicates the likelihood of each of the $K$ patterns occurred in each of the $N$ regions. $(W_{t})_{2N \times K}$ shows the probability of a latent pattern having a certain feature. Here, $K$ is the number of desired latent patterns to be found in the analysis. It's a hyperparameter usually given by analysts before the analysis.

\begin{figure}[!t]
    \centering
    \includegraphics[width=\linewidth]{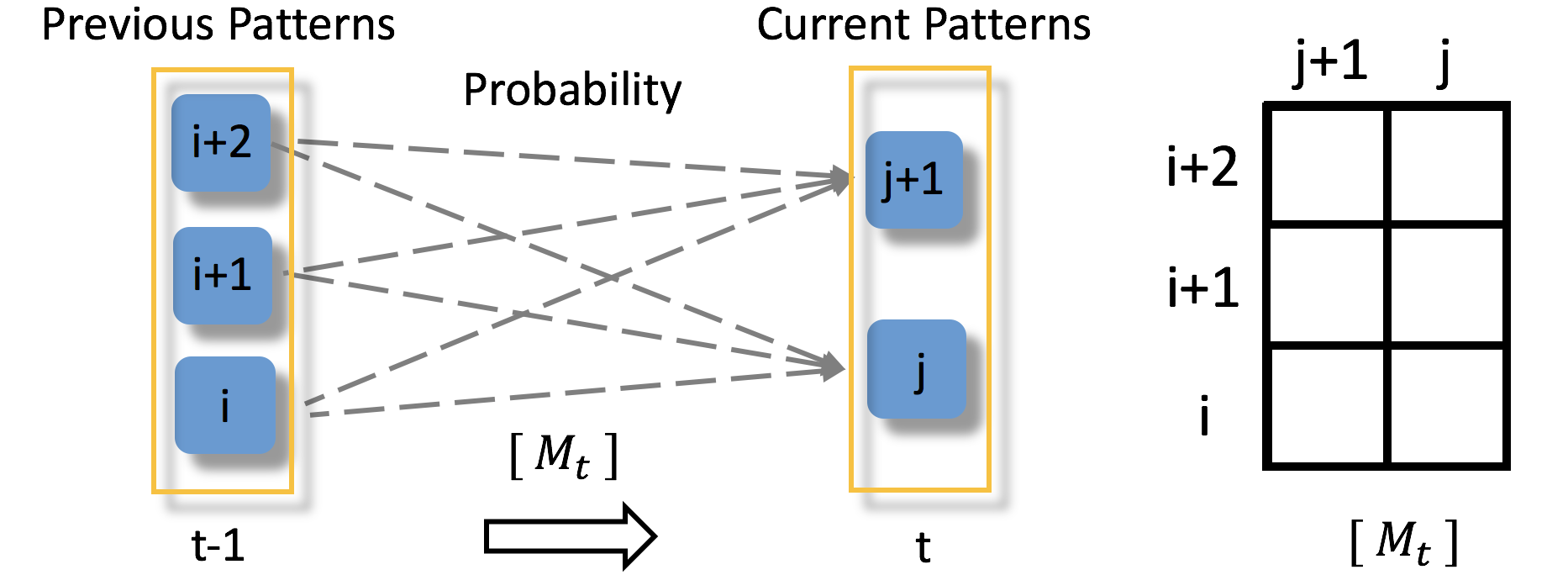}
    \caption{The blue rectangles in the left block indicate the previous patterns extracted at time interval $t-1$, while the blue rectangles in the right block indicate the current patterns at time interval $t$. $M_{t}$ is the transition matrix which implies the probabilities of the previous patterns at time interval $t-1$ transiting to the current patterns at time interval $t$.}
    \label{fig:transition}
\end{figure}

\textbf{Temporal Smoothness.} The second and the third terms are the regularization terms that ensure the temporal smoothness of the analysis result. They respectively preserve the similarity of $W_t$ and $H_t$ across different time to eliminate the dramatic sudden change due to the noisy data that may break the overall the transition trend of the regional mobility patterns. $\alpha$, $\beta$ controls the degree of smoothness. Considering the number of patterns $K$ vary from time to time, a transition matrix $M_t$ is introduced to connect patterns at different time intervals together. As shown in Fig.~\ref{fig:transition}, an element ${M_{t,i,j}}$ in $M_{t}$ implies the probability of a previous pattern $i$ at time $t-1$ transiting to a current pattern $j$ at time $t$.

\textbf{Spatial Smoothness.} The fourth term is the spatial smooth regularization term that ensures a region will share similar mobility patterns with its neighborhood. This design is due to the common understanding and observation that nearby regions will show similar mobility patterns as the functional area (e.g. CBD area) may locate across multiple adjacent regions. Here, an adjacency matrix $A$ is introduced with each element $A_{ij} \in \{1, 0\}$ indicates whether or not two regions $i$ and $j$ are adjacent to each other. The pattern differences (i.e., $||h_{t,i}-h_{t,j}||^{2}$) between those adjacent ones are minimized and the degree of the minimization is controlled by $\lambda$.

The above optimization problem can be solved based on block coordinate descent~\cite{kim2014algorithms}. The outputs of the algorithm, including the latent patterns $W_t$, the pattern distribution in different regions $H_t$, and the pattern transition probability $M_t$, are captured at each timestamp $t$. Based on $H_t$, we group grids sharing similar mobility patterns into latent regions via a clustering analysis. All these produced analysis results are used for building the meaningful visualization views that will be introduced in the next section.

\textbf{Discussions.} There are several important issues related to the above algorithm that are worth discussing.

\underline{\it Orthogonality}. Compared to singular value decomposition (SVD), one of the major limitations of NMF is that the orthogonality of its result is not guaranteed. To alleviate this problem and avoid extracting similar patterns which are hard to differentiate, a constraint $WW^{T}=I$~\cite{ding2006orthogonal} that ensures the orthogonality of the extracted patterns are introduced into the optimization process.

\underline{\it Choosing $K$}. The number of latent patterns, $K$, is usually the prior knowledge provided by the analysts. However, in many real applications, the ground truth of $K$ is unknown. To combat this issue, in our implementation, we employ Mean-Shift, a non-parametric clustering algorithm~\cite{cheng1995mean}, to compute the clusters at each timestamp and then use the numbers of clusters as the values of $K$ for our analysis. Although this approach is heuristic, it provides meaningful results (in Section~\ref{sec:eval}).

\underline{\it Comparing to SVD.} When compared to SVD, our algorithm benefits both in the computation efficiency and the ease of results interpretation. Decomposing an $M \times N$  matrix takes our algorithm $O(MNK)$ in time complexity compared to $O(MN^{2})$ of that of SVD. At most of the time, it is much more efficient as $K$, the number of patterns, is usually much smaller than $N$. In addition, unlike SVD which may produce negative values in the analysis results, our method guarantees non-negative values, which are meaningful and interpretable.


\section{Visualization}
\label{sec:visual}
In this section, we describe the visualization designs that illustrate the above analysis results. We begin with the design tasks, followed by the designs of detailed visualization views and interactions.

\subsection{Design Tasks}
We identified a set of visualization design tasks via several discussions with a domain expert and summary of the limitations of the existing techniques such as MobiSeg~\cite{wu2016mobiseg}. Specifically, we focused our design on how to efficiently reveal and display mobility patterns and dynamics of the latent regions over time.
\begin{figure*}[!htb]
    \centering
    \includegraphics[width=\linewidth]{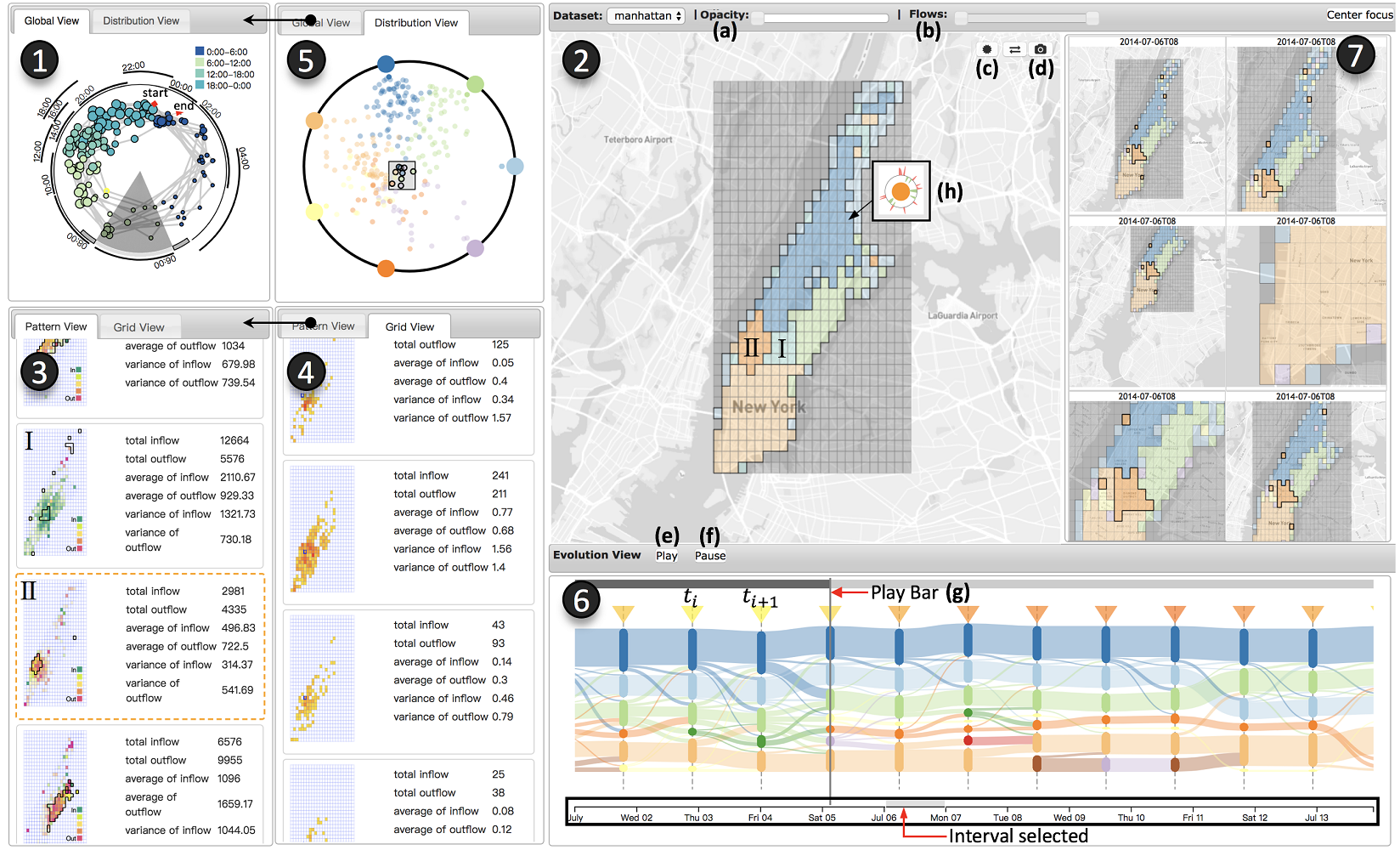}
    \caption{The user interface of \name system consists of seven major views: (1) global view; (2) map view; (3) pattern view; (4) grid view; (5) distribution view; (6) evolution view; and (7) snapshot view. 
    The interface also contains eight major components that support interaction: (a) opacity slider; (b) flow slider; (c) glyph button; (d) snapshot button; (e) play button; (f) pause button and (g) play bar.} 
	\label{fig:interface}
\end{figure*}
\begin{enumerate}

\item[{\bf T1}] {\bf Summarizing the change of the city over time.}  It is necessary to provide a high-level summarization of the overall change of the city over time so that analysts can quickly find out the dramatic changes for investigation.

\item[{\bf T2}] {\bf Showing the regional mobility patterns within the spatiotemporal context.}
To support the efficient identification and interpretation of the latent regions, the visualization should present the segmentation result of the latent regions with geographic information, the mobility features of the patterns (e.g., incoming and outgoing taxi trips within an area), and the evolution of latent regions within the temporal context. 

\item[{\bf T3}] {\bf Facilitating the analysis result validation and interpretation.} The visualization should reveal the raw traffic data inside each latent region and provide the corresponding statistical information for the interpretation and validation of the region segmentation results.

\item[{\bf T4}] {\bf Easy comparing of the regional patterns over space and time.} To support a comprehensive understanding of urban evolution, the visualization should enable the comparison of the mobility patterns or the segmentation results at different time intervals in different areas.
\end{enumerate}

\subsection{User Interface}
The design of the user interface is guided by the above design tasks. The interface consists of seven views (Fig.~\ref{fig:interface}): 
(1) the \textit{global view} that shows the overview of the segmentation results in the temporal context \textbf{(T1)};
(2) the \textit{map view} that combines the segmentation results with geographic information, as well as a flow glyph design that illustrates the raw traffic trip information of the regions \textbf{(T2)};
(3) the \textit{pattern view} and (4) the \textit{grid view} that present the mobility feature of each latent region, as well as statistics of the raw information to help validate the mobility pattern \textbf{(T2, T3)};
(5) the \textit{distribution view} that displays the probability distribution of different mobility patterns over regions \textbf{(T3)};
(6) the \textit{evolution view} that shows the dynamics of the latent regions in the temporal context \textbf{(T1, T2)} and (7) the \textit{snapshot view} that allows users to take a snapshot of the segmentation results shown in the map view for later retrieval and further analysis \textbf{(T4)}. 
These views are interactively linked to illustrate the dynamics of latent regions generated based on the aforementioned segmentation results. 

\textbf{Use case scenario.} 
We present the following scenario to help illustrate the potential usage of the \name system. Suppose Alice, an urban planner, attempts to explore and analyze the evolution of mobility patterns in an urban context using traffic data. 
The system provides an overview of the segmentation results in the global view (Fig.~\ref{fig:interface}(1)), through which Alice can compare the time intervals in terms of similar urban segmentation and selects the time intervals of interest. The corresponding information of the regions within the selected time intervals is updated and visualized in other views.
To observe what happened in these regions, Alice first analyze the latent regions with geographic information in the map view to understand the urban segmentation during this time interval(Fig.~\ref{fig:interface}(2)) and select the latent regions of interest to explore. The pattern view (Fig.~\ref{fig:interface}(3)) and grid view (Fig.~\ref{fig:interface}(4)) further show the features (i.e., incoming and outgoing traffic flows) of the selected regions and reveal the mobility patterns.
To validate the mobility pattern of these latent regions, Alice can click the glyph button (Fig.~\ref{fig:interface}(c)) to reveal the \textit{flow glyph} (Fig.~\ref{fig:interface}(h)) in the map view that presents the detail of the traffic trips.
She can also navigate the distribution view (Fig.~\ref{fig:interface}(5)) to analyze the detailed information about the probability distribution of different patterns over each region.
Now Alice gains an insight into the latent regions in the selected time interval and she intends to compare the segmentation results between different time intervals.
She jumps to the evolution view (Fig.~\ref{fig:interface}(6)) and drags the play bar (Fig.~\ref{fig:interface}(g)) to display the evolution of mobility patterns. She can observe the animated changes of segmentation in the map view. Once interesting areas are founded, she captures them in the snapshot view  (Fig.~\ref{fig:interface}(7)) for later review and comparison. 
Through these steps, Alice can understand the dynamics of latent regions and perform tasks for urban planning.

We employ four color encoding schemes in our design.
The first color scheme ranging from light green to dark blue shows the categorization of segmentation results in the global view (Fig.~\ref{fig:interface} (1)). 
The second color scheme is used to encode the segmentation results in the map view (Fig.~\ref{fig:interface} (2)), distribution view (Fig.~\ref{fig:interface} (5)), and evolution view (Fig.~\ref{fig:interface} (6)). Regions with the same color at a given time range indicates they share similar mobility patterns over spaces. Regions with the same color across different time range indicates these regions have a similar pattern changing trend.
The third color scheme is used to represent the difference between the incoming and outgoing traffic flow in the pattern view (Fig.~\ref{fig:interface} (3)) and the glyph in the map view. The colors ranging from green to yellow, and to red indicate the larger, equal, and smaller incoming flow when compared to the outgoing flow. 
We use the fourth color scheme ranging from yellow to red to indicate the amount of the flow in the grid view (Fig.~\ref{fig:interface} (4)) and the triangular glyph in the evolution view (Fig.~\ref{fig:interface} (6)).

\begin{figure}[tb]
    \centering
    \includegraphics[width=\linewidth]{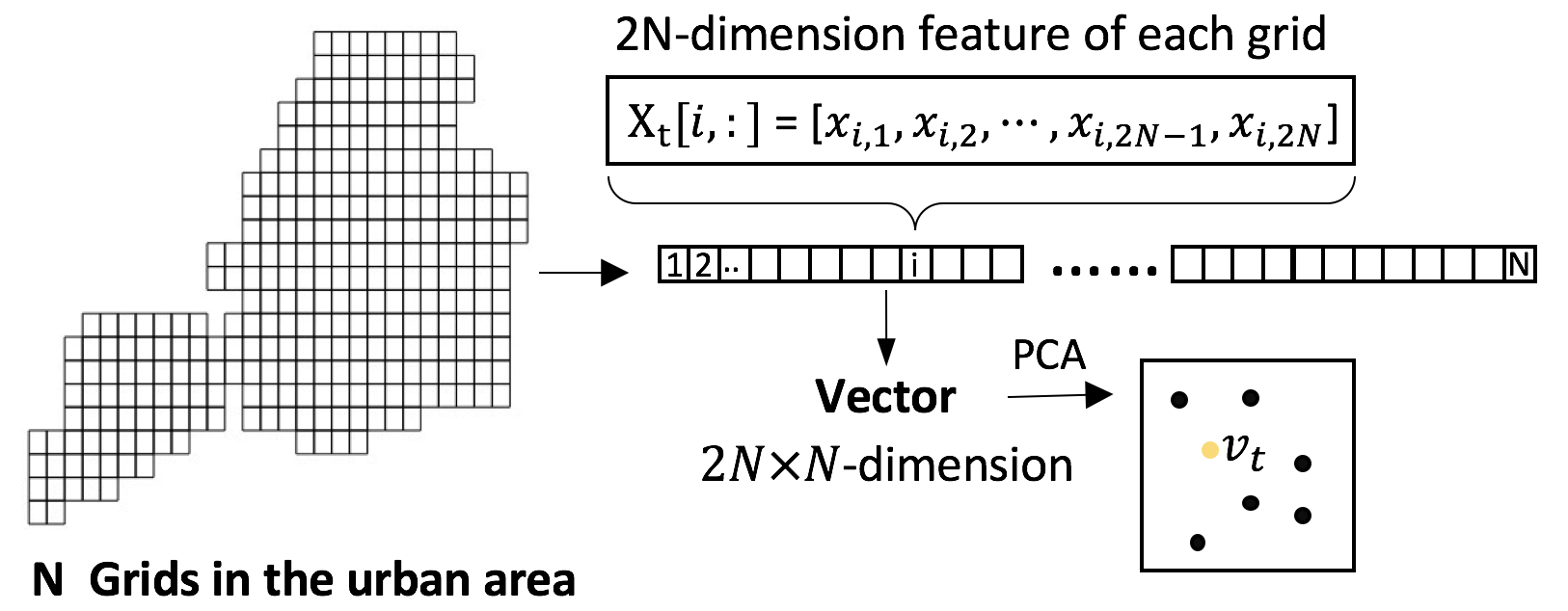}
    \caption{The characteristic of point $v_t$ is captured by an $N\times 2N$-dimensional feature vector, where $N$ indicates the number of grids in a region and $2N$ is the number of features of each grid. }
	\label{fig:grid}
\end{figure}

\subsection{Global View}
The global view (Fig.~\ref{fig:interface}(1)) provides an overview of the segmentation results in time series whose design is inspired by ~\cite{van2016reducing}. It illustrates the distribution of the overall segmentation results of a focal area at different time in a feature space. In particular, the segmentation results at different time are summarized and shown as points in the view with the point size indicating the amount of traffic within the corresponding time period and the color indicating the parts of the day (i.e., dawn, morning, afternoon, and night). The characteristic of a point, $v_t$, is captured by an $N\times 2N$-dimensional feature vector which is the vectorization results of the corresponding feature matrix $X_t$ introduced in Section~\ref{sec:algorithm} as shown in Fig.~\ref{fig:grid}. Here, $N$ indicates the number of grids and $2N$ is the number of features of each grid. 

With the above feature vector, we illustrate the distribution of the overall segmentation results (i.e. points in the view) in the feature space via principal component analysis (PCA)~\cite{dunteman1989principal}. These points are further connected by a timeline from the earliest time to the latest time in the data with the start and end points respectively marked with a red rectangle and red arrow. 

Fig.~\ref{fig:interface}(1) illustrates the visualization results of the aforementioned NYC taxi trip data collected from the Manhattan area, which forms a periodical circular pattern with each loop in the circle indicates a day that is segmented into 12 time periods. Within each period, a segmentation is calculated and thus are visualized as a point in the view. A circular brush tool is also designed in the view to facilitate the selection of a time range and the corresponding points. The selected points will be expended into details and shown in other visualization views for exploration and comparison. 

\subsection{Map View}
We overlay the segmentation analysis results on a map as shown in Fig.~\ref{fig:interface}(2) to illustrate its spatial context. In particular, the equal-sized grids, in which the mobility features are calculated, are visualized in the background. Each grid $i$ is colored by their primary mobility patterns (i.e., the largest field in the vector $H_t[i,:]$). The grids share similar mobility patterns are grouped together into latent regions and the boundary of the latent regions are further highlighted by a thicker line.

\begin{figure}[!htb]
    \centering
    \includegraphics[width=\linewidth]{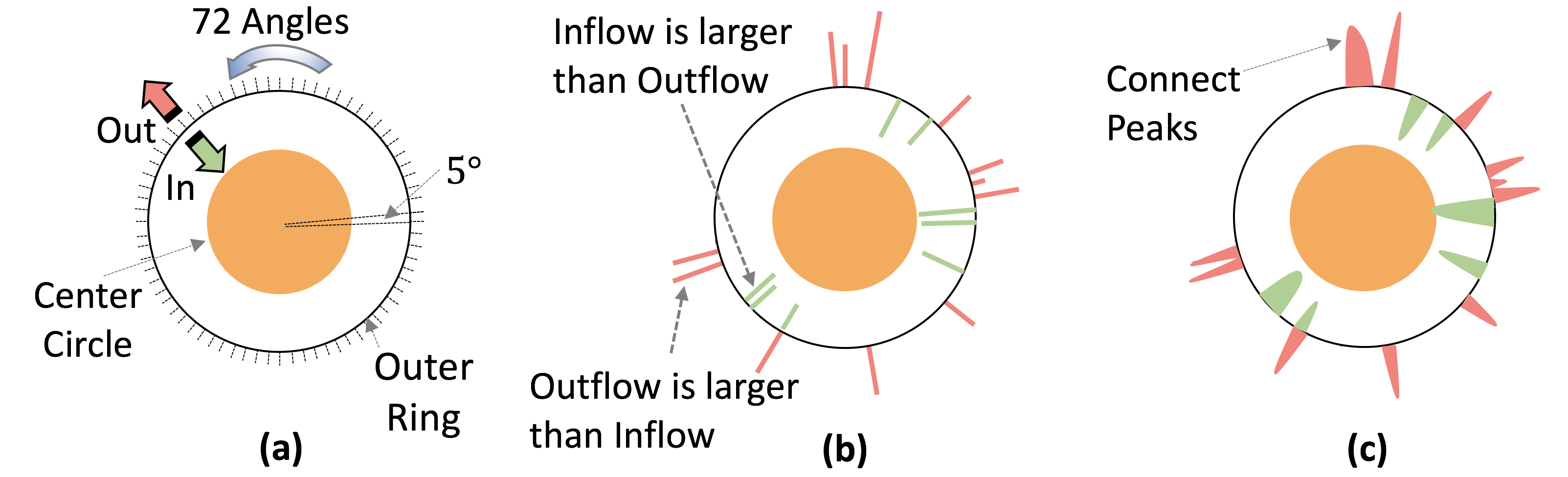}
    \caption{Flow-glyph design. (a) The center circle and outer ring respectively show the amount and direction of traffic flows in the grid. (b) These flows are visualized as bars and (c) enhanced by a colorful peak with red indicates outflow and green indicates inflow.}
	\label{fig:glyph}
\end{figure}

To summarize and illustrate the raw traffic flows inside each grid, we introduce a novel \textit{flow-glyph} design in the map view as shown in Fig.~\ref{fig:glyph}. The design of this glyph aims to encode and illustrate three types of the following information regarding to a focal grid: (1) the traffics inside the grid; (2) the exchange of the traffics between the focal grid and other grids; (3) the statistic of the total amount of traffics related to the grid. The glyph, as shown in Fig.~\ref{fig:glyph}(a), follows a circular design consists of two major components: (1) the center circle with the size indicating the total amount of relevant traffic flows in the grid, and (2) the outer ring that summarizes the traffics to or from 72 different directions (5 degrees a direction) with the focal grid in the center. These number of the flows are visualized as bars (Fig.~\ref{fig:glyph}(b)) and enhanced by a colorful peak with red indicates outflow and green indicates inflow (Fig.~\ref{fig:glyph}(c)). Intuitively the outflows are visualized outside the ring whereas the inflows are visualized inside the ring.

\subsection{Evolution View}
The evolution view illustrates the temporal transition trend of the latent regions as shown in Fig.~\ref{fig:interface}(6). It employs a Sankey diagram design in which x-axis indicates the time and the vertical nodes at each timestamp indicate the latent regions generated at that time with the node color consists to the color used in the above map view. The transitions of the latent regions across different timestamps are shown by the strips whose thickness indicates the number of raw grids merged into or split from a latent region. Here, the colors of the strips provide a strong visual hint, from which an analyzer can trace the change across different timestamp.

\begin{figure}[!htb]
    \centering
    \includegraphics[width=\linewidth]{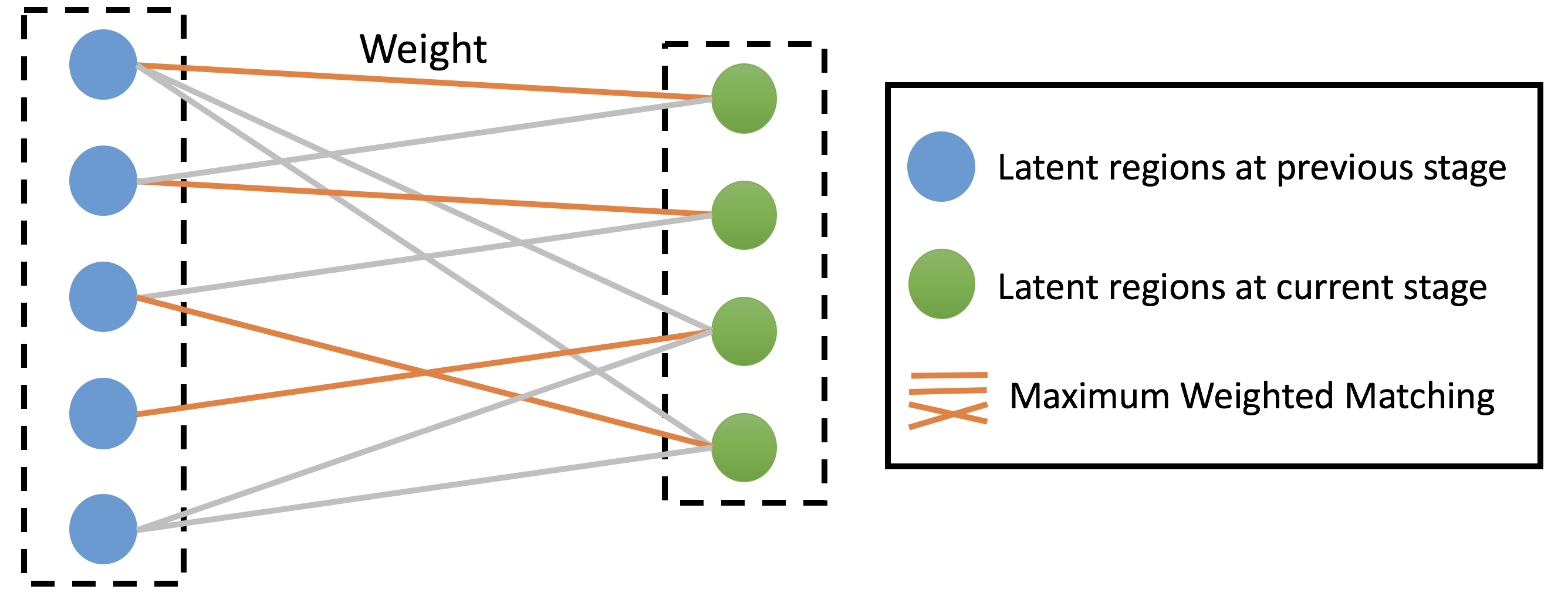}
    \caption{The color matching of the latent regions. The blue vertices are the latent regions at previous stage and the green vertices are the latent regions at current stage. A blue vertex and a green vertex are connected if they have overlapped grids and the weight of the connection is given by the number of the overlapped grids. The solution of the maximum weighted matching are links highlighted in orange.}
	\label{fig:KM}
\end{figure}

Sometimes, the number of latent regions may vary dramatically over time, thus making the assignment of a proper color to a node or a strip is a difficult problem. The goal is to find the best color matching so that the colors of the succeeding nodes and strips can best inherit from the colors of the previous nodes so that users can easily follow and track the transition trend of the latent regions. We convert the problem into a maximum-weighted bipartite matching problem by construction a bipartite graph. In particular, as shown in Fig.~\ref{fig:KM}, we connect a latent region at a previous stage (shown as a blue node) to a latent regions at the current stage (shown as a green node) if these two regions have overlapped underlying grids and the weight of the connection is given by the number of the overlapped grids. In this way, a maximum-weighted matching between the blue nodes and the green nodes will help find a way for the green nodes best inherent colors from their most relevant nodes at the previous stage. The problem can be optimally solved based on the Kuhn-Munkres algorithm~\cite{kuhn1955hungarian,munkres1957algorithms}. It is worth mentioning that in our implementation based on the NYC taxi trip data, we initially segment the Manhattan into functional regions and assign each region a color based on the administrative divisions of the city. These initial colors are then used for the color assignments and matching during the rest of regional transition processes.

\subsection{Other Views}

The \name also support others views to illustrate information details from different perspectives.

\textbf{Pattern View.}
The pattern view (Fig.~\ref{fig:interface}(3)) shows a list of mobility patterns for each latent region.
As described in Section~\ref{sec:algorithm}, the mobility pattern of a latent region can be presented by a 2N-dimensional feature vector (a column vector in the matrix $W_{t}$).
The first $N$ fields in the vector indicate the occurrence probability of the outgoing trips from this latent region to all the $N$ grids in the urban area, while the last $N$ fields indicate the occurrence probability of the incoming trips from all the grids to this latent region.

To visualize this feature vector intuitively, we employ a red-to-yellow-to-green color gradient on the heatmap of each latent region. 
If the probability that people in a grid enter into the latent region (highlighted via black strokes in the heatmap) is higher than the probability that the people leave from the latent region to this grid, the grid in the heatmap will be colored in green. Otherwise, the grid will be colored in red. 

For example, in the heatmap of the highlighted item (Fig.~\ref{fig:interface}(3-II)), most of the grids on the east side of the latent region are colored in red. This pattern suggests that people in latent region \uppercase\expandafter{\romannumeral2} are likely to enter into the east side of this latent region. In the map view we can observe that the east of  latent region \uppercase\expandafter{\romannumeral2} is latent region \uppercase\expandafter{\romannumeral1}. 
We further inspect the mobility pattern of latent region \uppercase\expandafter{\romannumeral1} (Fig.~\ref{fig:interface}(3-I)) and find that the grids around latent region \uppercase\expandafter{\romannumeral1} are colored in green, indicating people in those grids are likely to enter into this latent region.
Therefore, we could infer that people in latent region \uppercase\expandafter{\romannumeral2} intend to travel to latent region~\uppercase\expandafter{\romannumeral1}.

\textbf{Grid View.} 
As shown in Fig.~\ref{fig:interface}(4), the grid view further reveals the detailed mobility patterns of grids that compose a certain latent region selected in the pattern view. Each item in the grid view consists of statistic information and a heatmap. In the heatmap, the color encodes the total amount of flow between the grid and others. The darker the color, the larger the amount.

\textbf{Distribution View.}
The distribution view (Fig.~\ref{fig:interface}(5)) reveals relationship among mobility patterns and the regions by visualizing the details of the matrix $H_{t}$. Each row vector of the matrix $H_{t}$ indicates the probability distribution of the mobility patterns over a certain region.

A circle is divided into $K$ (the number of the mobility patterns derived from Section~\ref{sec:algorithm}) sections equally, so as to build a barycentric coordinate. The categories of the mobility patterns are illustrated as colored nodes along the boundary of the circle. The regions are encoded as scattered points in the barycentric coordinate. Each pattern and its corresponding points of the regions are assigned with a specific color. The coordinate of the point representing region $i$ is calculated as~follows:
\begin{align}
rx_{i}=\sum_{j}^{k}H_{t,i,j}\times cx_{j},~~ry_{i}=\sum_{j}^{k}H_{t,i,j}\times cy_{j}
\label{eq:dis}
\end{align}
where $rx_{i}$ and $ry_{i}$ are x-coordinate and y-coordinate of region $i$, $cx_{j}$ and $cy_{j}$ are x-coordinate and y-coordinate of mobility pattern $j$, and $H_{t,i,j}$ is the element at row $i$ and column $j$ of matrix $H_{t}$.

\subsection{Interactions}
The system provides rich interactions to help users analyze mobility patterns from different perspectives and obtain a comprehensive understanding of urban evolution.

\textbf{Filter.} 
In the map view (Fig.~\ref{fig:interface}(2)), two sliders are designed to enable users to manually filter irrelevant information and thus focus on the regions of interest. 
The opacity slider (Fig.~\ref{fig:interface}(a)) is used to set the opacity of each region. High opacity reveals more details of the underlying geographic map for the region, while low opacity makes it easier for users to distinguish between different latent regions. 
The flow slider (Fig.~\ref{fig:interface}(b)) is a two-way slider for selecting the range of traffic volume of the latent regions. It helps filter out regions with the total flow amount within the selected range, while others are greyed out.

\textbf{Select Time Interval.} 
Users can select a certain time interval by clicking the triangle glyph on top of the time-stamp axis in the evolution view or clicking the points in the global view. 

\textbf{Link.}
The system supports automatic linking among the proposed visualization views. 
When users hover the mouse over a grid in the map view or a point in the distribution view, the corresponding latent region in the map view will be highlighted with black strokes. 
When selecting a time interval in the evolution view, the corresponding point in the global view will be highlighted.
In addition, when hovering over a node along the circle in the distribution view or over the items in the pattern view, the corresponding mobility pattern in the evolution view (i.e., a node at a timestamp) will be highlighted. 

\textbf{Brush and Zoom.} 
The system enables zooming and panning for exploring the Sankey diagram in the evolution view. Users can further brush on the timeline to select a certain time period for in-depth analysis. In addition, the system also enables brush in the distribution view for users to highlight a group of scattered points.

\textbf{Explore the Map.} 
In the map view, users can drag to focus on a certain urban area and zoom in for more geographic details. 
In addition, when users zooming in/out the map, the size of the flow glyph changes accordingly. In particular, when zooming out, although details of the glyph cannot be clearly shown, users can get an overview of the mobility pattern in the corresponding region by observing the shape of each glyph. 
In addition, when users click a certain item in the grid view, the viewpoint of the map view will smoothly transit to the location of the selected region through animation.

\textbf{Snapshot.} 
The system allows users to take a snapshot of the map view to facilitate comparisons of the results at different time intervals. 
By clicking the snapshot button (Fig.~\ref{fig:interface}(d)), users can capture the current map view into the snapshot view.
Then users just need to click it in the snapshot view to recover a snapshot in the map view whenever it's~needed.

\textbf{Adjust.} 
In the map view, users are allowed to modify the results to better align with their domain knowledge. Users can simply right-click on a specific grid in a region to change its color. 
This feedback can help refine the overall result, and the map view will update accordingly by presenting a new segmentation result.

\textbf{Auto-play.}  
The evolution view supports automatic scan along the timeline.  
When users click on the play button (Fig.~\ref{fig:interface}(e)), the play bar (Fig.~\ref{fig:interface}(g)) will scan over the timeline and the segmentation results in the map view will be updated continuously. 
Clicking the pause button (Fig.~\ref{fig:interface}(f)) will stop this procedure and users can further analyze the result of a certain time interval. 

\textbf{Switch Context.}
The system enables users to explore rich context information through interactions.
Users can switch between different views (i.e., the global view and the distribution view, the pattern view and the grid view) by clicking the corresponding tabs. In addition, users can switch on glyphs of grids in the map view to inspect the traffic flow in each region by clicking the glyph button~(Fig.~\ref{fig:interface}(c)). 

\begin{figure}[!b]
    \centering
    \includegraphics[width=\linewidth]{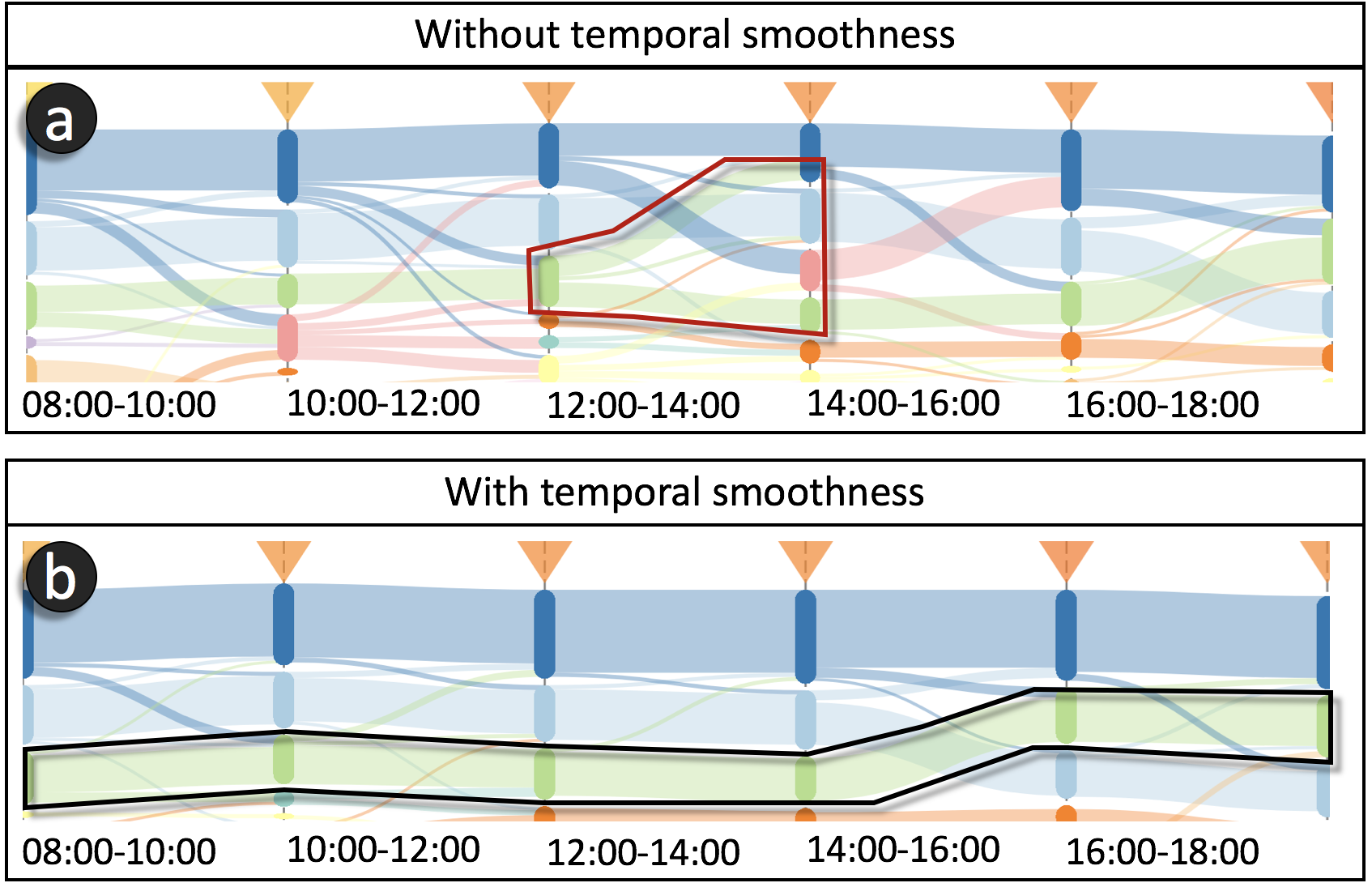}
    \caption{The analysis results in the evolution view generated (a) without and (b) with the temporal smooth regularization terms in the algorithm.}
	\label{fig:temporal}
\end{figure}

\section{Evaluation}
\label{sec:eval}
In this section, we evaluate the effectiveness of the proposed algorithm and the \name system via comparative analysis, case studies and a domain expert interview. Our evaluation is based on a dataset consisting of over 450,000 taxi trips during July 2014 to December 2014 collected from over 50000 Yellow Cabs in Manhattan area. We segment the city into a web of grids with a granularity of $0.005 \text{ lng}\times 0.005 \text{ lat}$ and calculates the regional features every two-hour to ensures a reasonable computation time and a precise preservation of the traffic patterns.

\subsection{Algorithm Validation}
We validate the effectiveness of the algorithm by estimating the constraints and regularization terms introduced in our dynamic city segmentation algorithm.

\subsubsection{Verification of the Temporal Smoothness}
\begin{figure}[!htb]
    \centering
    \includegraphics[width=\linewidth]{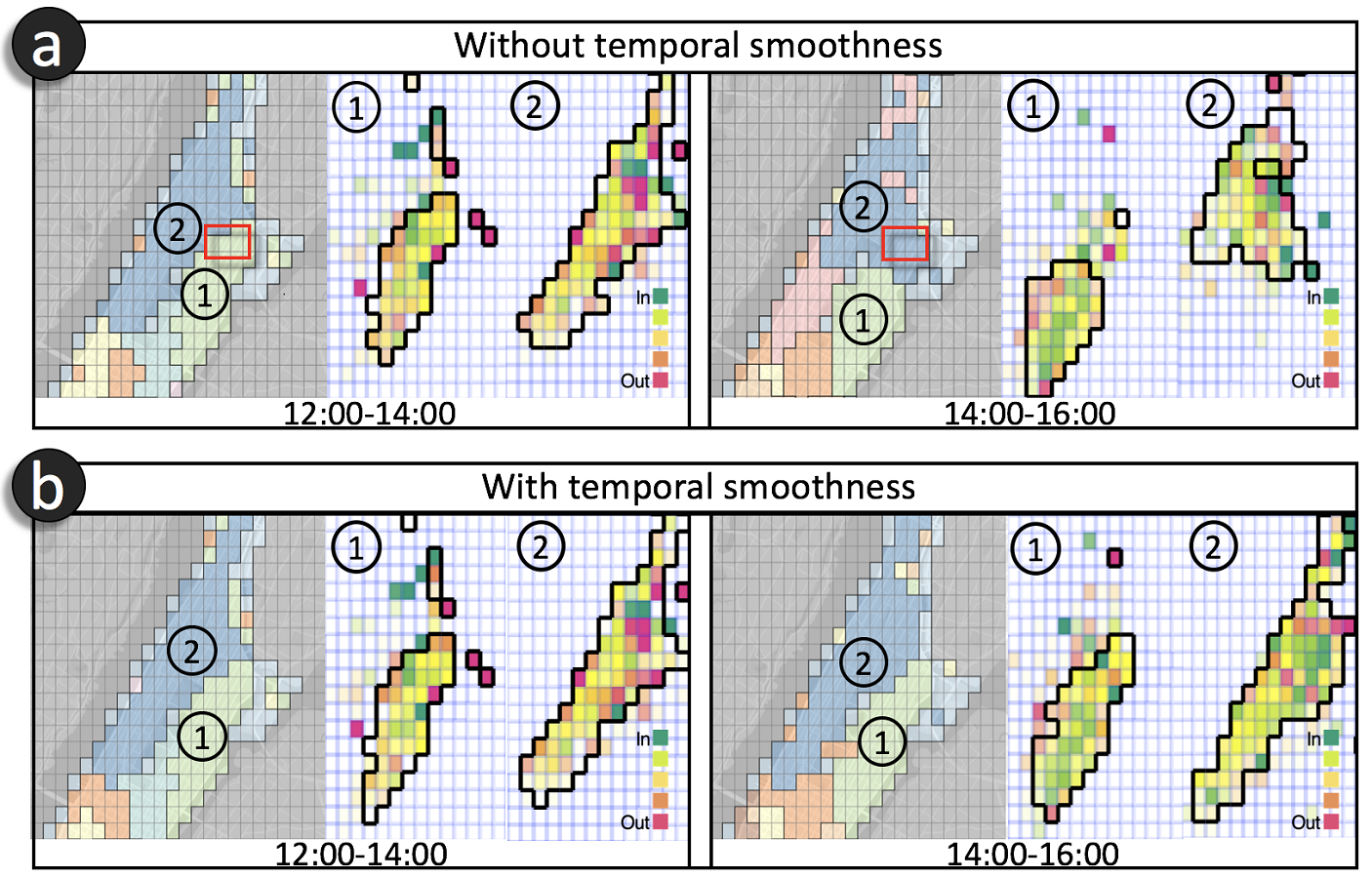}
    \caption{The analysis results in the map view and pattern view generated (a) without and (b) with the temporal smooth regularization terms in the algorithm.}
	\label{fig:temporal_detail}
\end{figure}

To verify the effectiveness of the temporal smooth regularization terms (Eq~\ref{eq:OP}(2,3)), we compare the analysis results generated without/with the term in the algorithm as respectively shown in Fig.~\ref{fig:temporal}(a) and (b). Generally, the transitions of the latent regions change dramatically in Fig.~\ref{fig:temporal}(a) when compared to the case shown in Fig.~\ref{fig:temporal}(b). In particular, the highlighted green strip (i.e., a latent region) in Fig.~\ref{fig:temporal}(a) splits into branches during the period from 12:00 to 14:00, which are later merged into another latent region shown as the blue strip in the next stage during the period of 14:00 to 16:00. The corresponding map view (Fig.~\ref{fig:temporal_detail}) provides more insights into the changes of these latent regions that helped with the results validation. In particular, blue and green regions respectively correspond to the above blue and green strips, illustrating the areas with two different latent patterns. These two areas changed dramatically in Fig.~\ref{fig:temporal_detail}(a): a subarea, highlighted by the red box, originally in the green region merged into the blue region. This subarea, according to the map, is the East Harlem, where schools, residential areas, and parks are located and the traffic patterns are seldom changed during the non-traffic hours like the period from 12:00 to 16:00.  A further investigation of the raw data verified our guessing, the change is due to a small number of random taxi trips which are the data noise that affects the analysis result. A slight smooth over temporal dimension addressed this problem as shown Fig.~\ref{fig:temporal_detail}(b), which verified the usefulness of the temporal smooth regularization term.

\subsubsection{Verification of the Spatial Smoothness}
\begin{figure}[!htb]
    \centering
    \includegraphics[width=\linewidth]{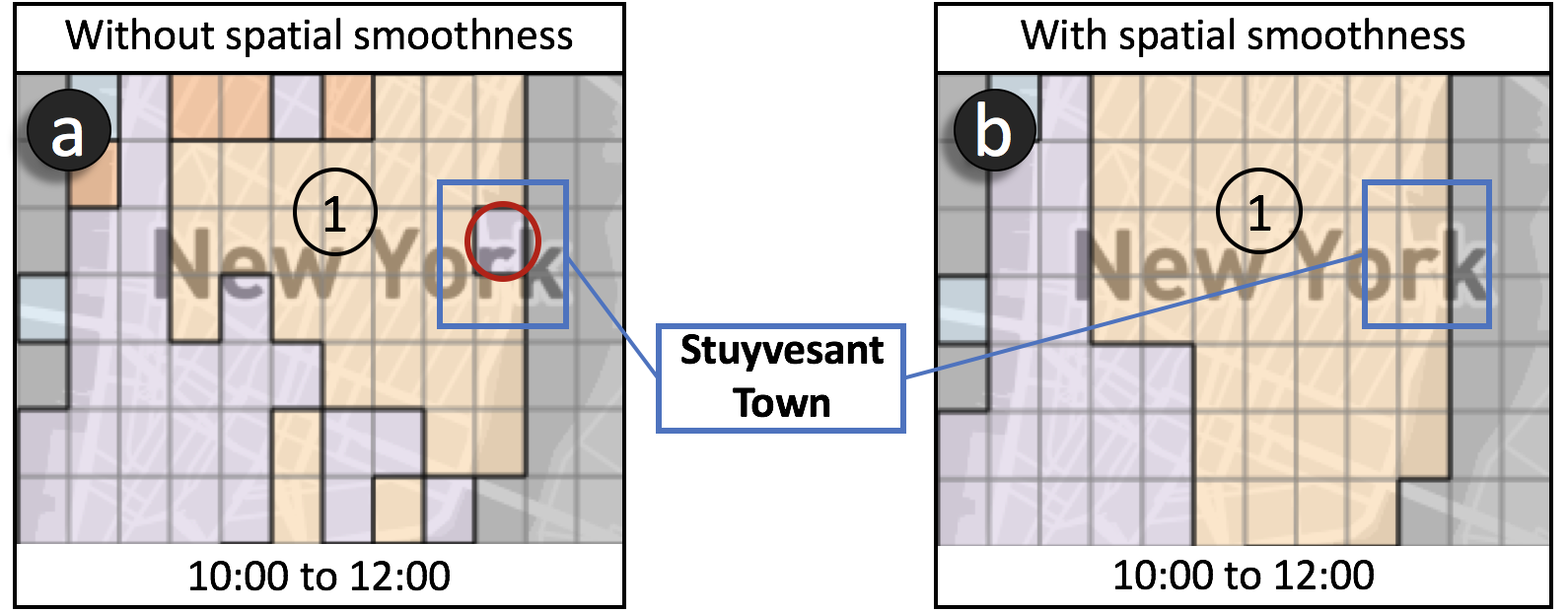}
    \caption{The analysis results in the map view generated (a) without and (b) with the spatial smooth regularization terms in the~algorithm.} 
	\label{fig:spatial}
\end{figure}

To verify the effects of the regularization term for spatial smoothing (Eq~\ref{eq:OP}(4)), we compare the analysis results produced without/with spatial smoothness as shown in Fig.~\ref{fig:spatial}(a) and (b) respectively. In this example, the region highlighted in the red circle is a part of Stuyvesant Town, a small residential area, in which people suppose to behave similarly. Therefore, the discontinuity of the regional clusters shown in Fig.~\ref{fig:spatial}(a) is most likely due to the noise of the input data instead of different mobility patterns. This problem has been eliminated by adding the spatial smoothness regularization term into our algorithm as shown in Fig.~\ref{fig:spatial}(b).

\subsubsection{Verification of the Orthogonality} 
\begin{figure}[!htb]
    \centering
    \includegraphics[width=\linewidth]{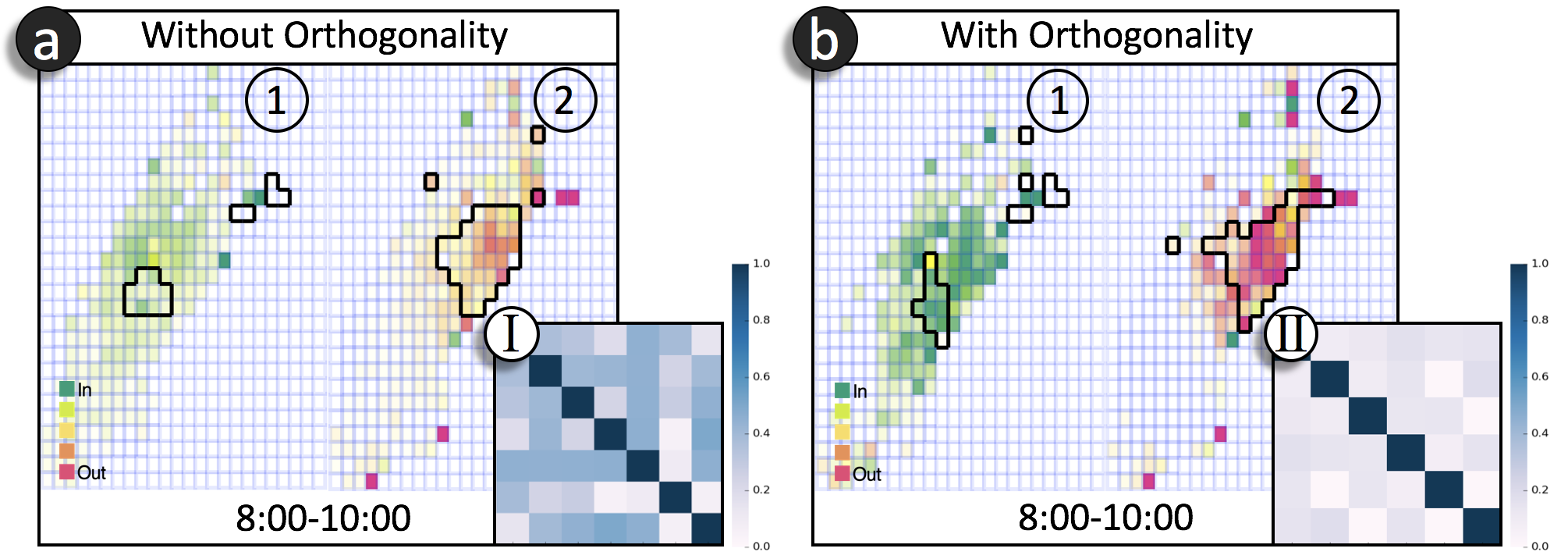}
    \caption{The analysis results in the pattern view generated (a) without and (b) with orthogonality constraint. (I) and (II) show the corresponding correlation matrices of (a) and (b) respectively.}
	\label{fig:orthogonal}
\end{figure}

To estimate the effects of the orthogonality constraint, we also compare the analysis results produced without/with the constraint as shown in Fig.~\ref{fig:orthogonal}(a) and Fig.~\ref{fig:orthogonal}(b) respectively. Obviously, the patterns shown in Fig.~\ref{fig:orthogonal}(b) are more differentiable than the one shown in Fig.~\ref{fig:orthogonal}(a). This finding is further verified by the corresponding correlation matrices shown in Fig.~\ref{fig:orthogonal}(I,II). In these matrices, each column or row indicates a latent pattern. A cell at the $i$-th row and the $j$-th column indicates the correlation value of the $i$-th and the $j$-th patterns, which is proportional to the color saturation. Therefore, Fig.~\ref{fig:orthogonal}(II) illustrates the patterns produced by following the orthogonality constraint are less relevant to each other when compared to the case shown in Fig.~\ref{fig:orthogonal}(I).

\subsection{Case Study}
To further evaluate the usability and usefulness of the \name system, we provide case studies demonstrating its capability in analyzing the change of the mobility patterns in Manhattan, NYC.

\begin{figure*}
  \centering
  \includegraphics[width=\textwidth]{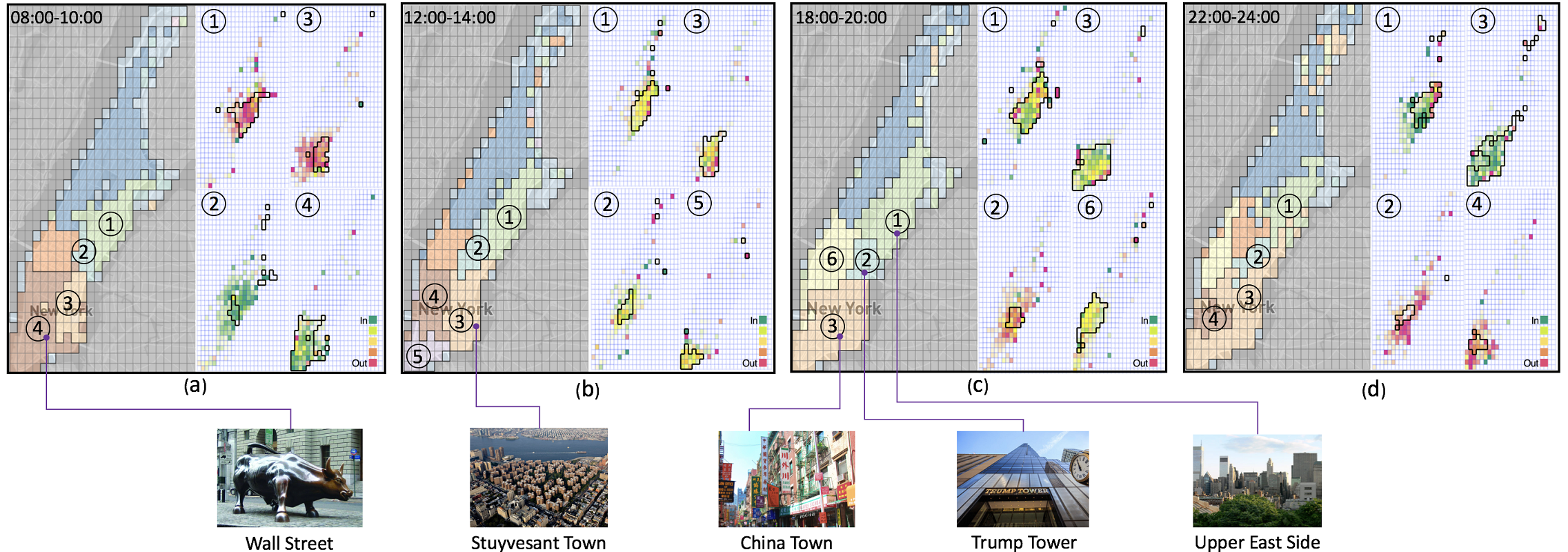}
  \caption{The \name system summarizes the daily evolution of the mobility patterns in Manhattan, NYC. 
  The system employs \textit{map view} (left) and \textit{region view} (right) to respectively show the result of urban segmentation and the mobility feature of each latent region.
  Four mobility patterns at different time intervals are captured, including (a) the morning from 8:00 to 10:00, (b) the early afternoon from 12:00 to 14:00, (c) the evening from 18:00 to 20:00, and (d) the night from 22:00 to 24:00.}
  \label{fig:teaser}
\end{figure*}

\subsubsection{Evolution Exploration}
We analyzed the daily evolution of the mobility patterns by comparing the results produced at four different time on $9^{th}$ July 2014, respectively in the morning, the afternoon, the evening, and at night. The visualization results are captured in Fig.~\ref{fig:teaser}.

As shown in Fig.~\ref{fig:teaser}(a), most of the grids in the latent region (1) and (3), the residential areas, are in red. This suggests that these areas have a greater outflow in the morning. In comparison, the regions marked as (2) and (4) have a greater inflow (shown in green) at the same time, where are actually the CBD (i.e. Central Business District) areas in the town. This reveals the mobility pattern of morning traffic hours within Manhattan.

The regions kept evolving. At noon and in the early afternoon (12:00 - 14:00), as shown in Fig.~\ref{fig:teaser}(b), the boundary of the latent regions (1,2,3) largely remain the same as that in the morning. However, the traffic patterns are dramatically changed as illustrated in the aside heatmap. The yellow color indicates the amount of incoming and outgoing traffic flows in these regions are similar, which implies people travel around for lunch inside the nearby regions. We also observed a new region (5), the financial area in the city, split from region (4). The aside heatmap revealed the reason for this change as the traffics are seldom across these two regions in this period of time, thus making them separated from each other. We believe people work in this area such as the stock dealer will be too busy to leave this area at noon.

Later in the evening (18:00 - 20:00) as shown in Fig.~\ref{fig:teaser}(c), the region (3, 4, 5) are merged together, forming an area labeled as (3) where people are moving around and the incoming and outgoing traffic flow are balanced. This is due to some people leaving for home and some of them coming for dinner and fun, since the area is also served as the entertainment district in the urban area where Broadway and many bars and restaurants are located. Similar patterns also occur in region (1) and (6). However, the traffic pattern shown in region (2) changes dramatically when compared to that of an early stage shown in Fig.~\ref{fig:teaser}(b). It suggests that people start to leave this area and the traffic spread all over the nearby regions, which is just opposite to the patterns shown in Fig.~\ref{fig:teaser}(a). 

Finally, Fig.~\ref{fig:teaser}(d) illustrates the traffic pattern from the CBD areas to the residential areas in the city, which is just opposite to the patterns shown in Fig.~\ref{fig:teaser}(a). 

\begin{figure}[!htb]
    \centering
    \includegraphics[width=\linewidth]{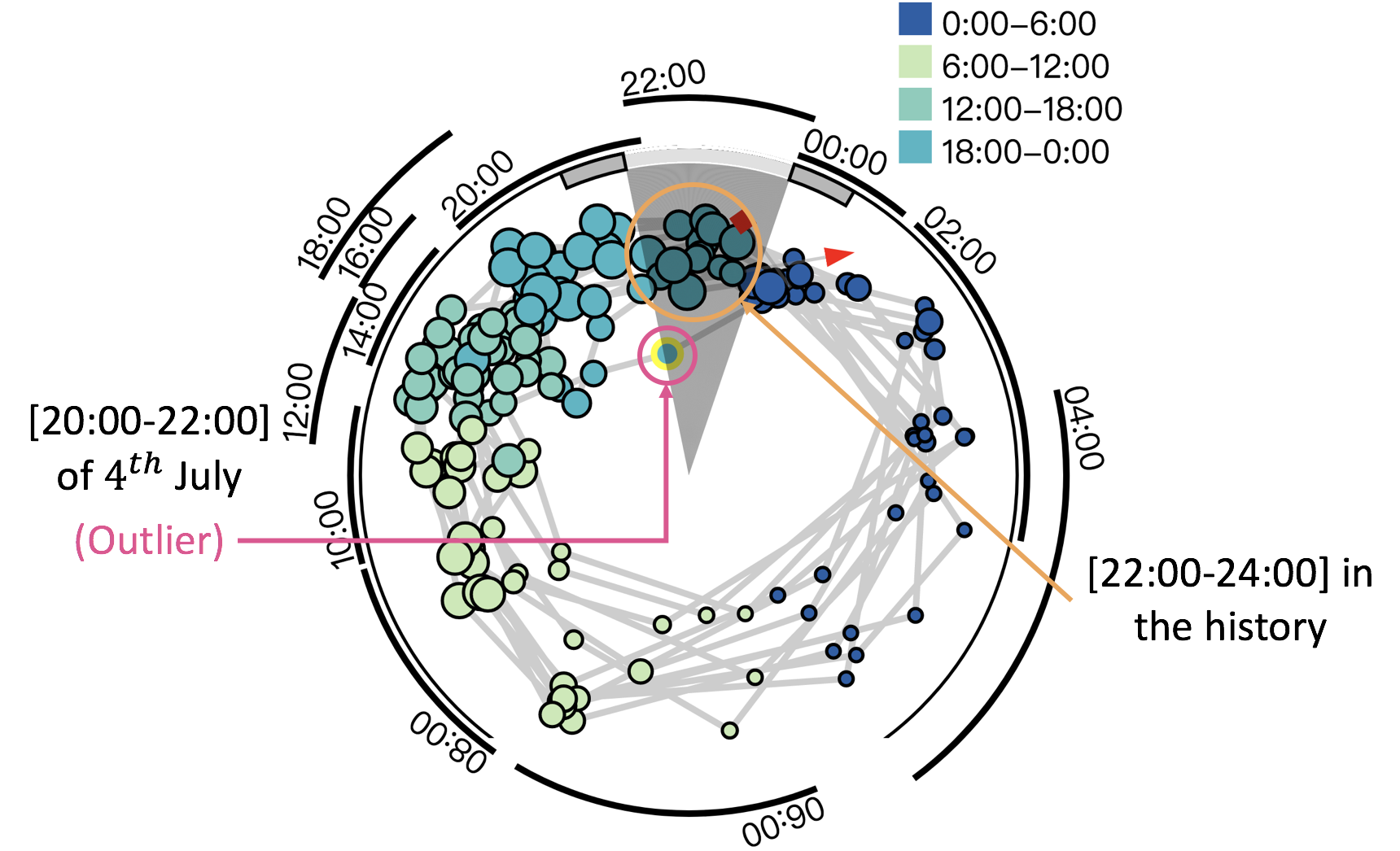}
    \caption{In the global view, a point representing the time interval 22:00~-~24:00 on July 4th (highlighted in the red circle) is considered to be an~outlier.} 
	\label{fig:global}
\end{figure}

\subsubsection{Anomaly Detection} 
\begin{figure}[!htb]
    \centering
    \includegraphics[width=\linewidth]{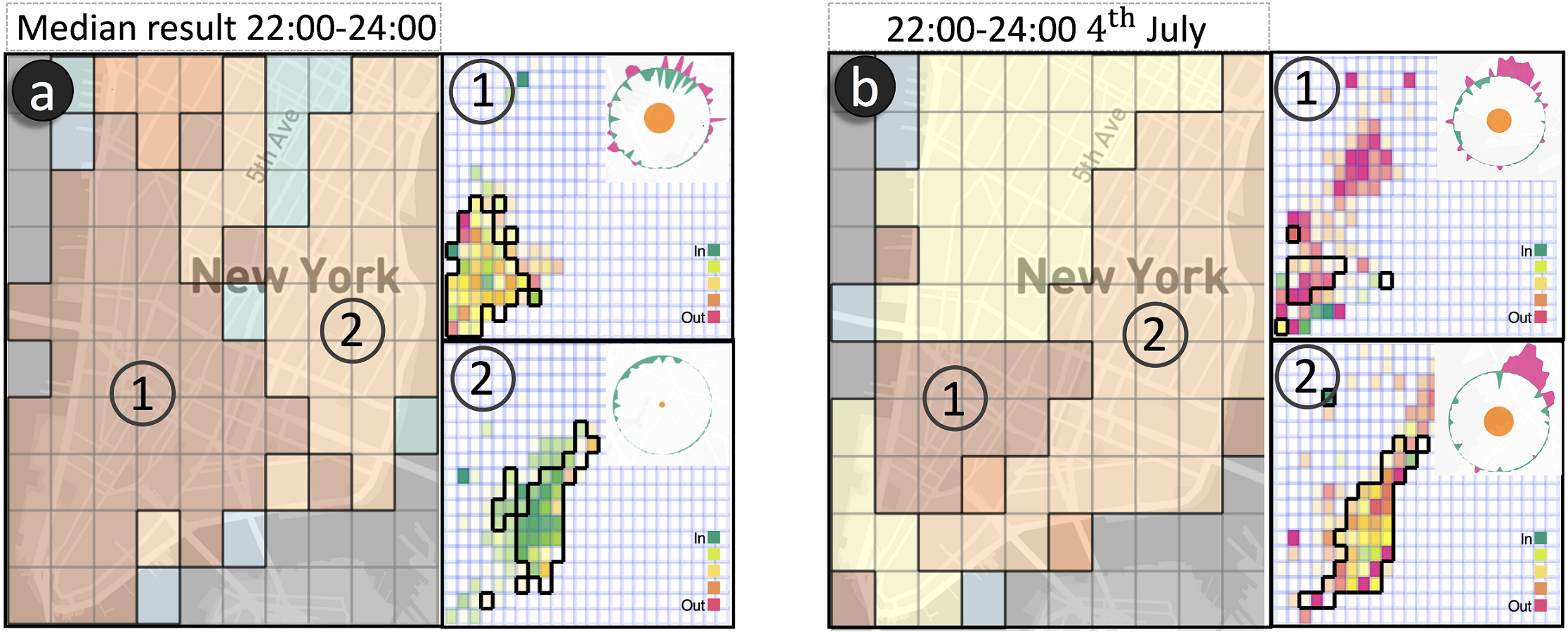}
    \caption{By comparing to (a) the normal pattern, (b) the anomalies traffic flows near the Brooklyn Bridge on July 4th, 2014 (the Independent day) is revealed.}
	\label{fig:independent}
\end{figure}

\name was also used to detect anomalous situations. Fig.~\ref{fig:global} illustrates an overview of the mobility patterns at different timestamp calculated based on people's daily moving behavior. It generally reveals a strong periodical pattern through the connected points shown in a circular form, with each circle indicates the period of a regular weekday. Among all the points connected by the timeline, a point representing the time interval 22:00-24:00 on July 4th, highlighted in the red circle, is considered to be an outlier as it is laid out away from other points in the surrounding context that captures the history of the same period of time. A detailed investigation on this outlier is illustrated in Fig.~\ref{fig:independent}. We believe this abnormal pattern is due to the change of traffics around the Brooklyn bridge on July 4th (the Independent day) during the period of 22:00-24:00. Usually, little volume of traffic goes through the bridge at the late night (Fig.~\ref{fig:independent}(a)). However, on the Independent day the fireworks aside the river attract a great number of people which dramatically increases the amount of traffic flow (Fig.~\ref{fig:independent}(b)), thus resulting in a different traffic pattern and captured by the global view of \name system. 

\subsection{Expert Interview}
We collected user feedback and comments on the \name system through an in-depth interview with a domain expert from the institute of urban planning and design in China. 
The interview lasted about 40 minutes.
We first showed a tutorial that explains the goal, visual encoding, and interactions of the \name system followed by a demonstration of an example illustrating the evolution process of an urban area to help the expert get familiar with the system. The expert was asked to explore the capabilities of our system and analyze the dynamic patterns in the urban area with his domain knowledge. 
During the process, we recorded the interview and took notes of his comments, which were summarized into four aspects:

\textbf{System:} Generally, the system impressed the expert. He commented that the results were ``reasonable'', and the design of the visualization views was ``intuitive'' and ``comprehensive''. ``We used to analyze the urban areas based on the statistical information overlaid on top of a map. This system provides us a new approach to exploring the dynamics of the urban data.'', which was considered to be ``novel" and ``useful". 

\textbf{Visualization:} The expert believed that most of the visualizations well supported the design requirements. For example, when analyzing the evolution of the urban area, the expert compared the results with the urban planning map and pointed out that ``the areas gather together in different time intervals showing different patterns. These patterns are meaningful given the functionality (e.g., residential or CBD areas) of the underlying urban areas.'' He also believed that the overview is useful as it provided a "clear periodical pattern" and "revealed a few outliers that fail to align with others". The expert also commented on the glyph design and believed it was a ``good and novel" approach for summarizing and illustrating raw trip data. He also believed the snapshot view was ``particular useful" for analysis and comparison tasks. At the last he said ``all these views aligned together are very comprehensive and provide useful information" and ``the system is quite useful in terms of supporting the exploration of city dynamics". 

\textbf{Interaction:} In terms of the interactions, the expert believed they were ``easy to use". He particularly like the auto-play function supported in our system, he felt these interactions were ``very nice", ``provide a flexible way for me to investigate temporal dynamics", and ``it is fun to play and watch the changes of the city" but he also pointed out that ``The animation is nice to watch but difficult to capture the change" and therefore, believed the snapshot function to be ``very useful for making a comparison".  

\textbf{Application:} The experts further mentioned the applications of this system in the field of smart-city services. ``With the help of this system, we can know more about the dynamics of the city. The information is useful in various applications, such as smart traffic light control for energy saving''. He also suggested that ``you can use different types of the mobility data, such as subway records and mobile phone locations, to detect more meaningful features for analyzing. [...] Currently, the analysis in each time period is static. Adding the time dimension to data matrix will make the result more powerful.''

\section{Conclusion}
\label{sec:conclusion}
In this paper, we present a visual analysis system, \name, for analyzing the dynamic latent regions that shape the urban area. 
\name is designed according to real-world requirements, such as interpreting mobility pattern and urban evolution. An NMF-based algorithm is introduced to reveal the regional dynamics of the mobility evolution patterns inside a focal urban area. We evaluate the performance and effectiveness of \name using a taxi-trip dataset of Manhattan Island through case studies and an interview with a domain expert. Our study results indicate our system is capable of identifying the mobility patterns to form latent regions, uncovering the dynamics of latent regions, and interpreting the mobility patterns of regions.
In the future work, we plan to use more data resources in the analysis and process the growing scale of data for real-time analysis.


\section{Acknowledgement}
Nan Cao and Yingcai Wu are the corresponding authors. We would like to thank all experts for participating our case studies and interviews. This research was supported in part by the National Natural Science Foundation of China Grants 61602306, Fundamental Research Funds for the
Central Universities, and the National Grants for the Thousand Young Talents in China.

\bibliographystyle{abbrv-doi-hyperref-narrow}

\clearpage

\end{document}